\documentclass{article}
\usepackage{pdflscape}
\usepackage{times}
\usepackage{graphicx} 
\usepackage{subfigure} 

\usepackage{amsmath}

\usepackage{natbib}

\usepackage{algorithm}
\usepackage{algorithmic}

\usepackage{hyperref}
\usepackage{array}
\newcolumntype{L}[1]{>{\raggedright\let\newline\\\arraybackslash\hspace{0pt}}p{#1}}
\newcolumntype{C}[1]{>{\centering\let\newline\\\arraybackslash\hspace{0pt}}p{#1}}
\newcolumntype{R}[1]{>{\raggedleft\let\newline\\\arraybackslash\hspace{0pt}}p{#1}}

\usepackage[accepted]{icml2017}

\icmltitlerunning{The Game of Tetris in Machine Learning}

\begin{document} 

\twocolumn[
\icmltitle{The Game of Tetris in Machine Learning}



\icmlsetsymbol{equal}{*}

\begin{icmlauthorlist}
\icmlauthor{Sim\'on Algorta}{to}
\icmlauthor{\"{O}zg\"{u}r \c{S}im\c{s}ek}{ed}
\end{icmlauthorlist}

\icmlaffiliation{to}{ABC Research Group at the Max Planck Institute for Human Development, Berlin, Germany}
\icmlaffiliation{ed}{University of Bath, Bath, United Kingdom}

\icmlcorrespondingauthor{Sim\'on Algorta}{algorta@mpib-berlin.mpg.de}

\icmlkeywords{reinforcement learning, tetris, sequential decision making under uncertainty, genetic algorithms, videogames}

\vskip 0.3in
]



\printAffiliationsAndNotice{}  

\begin{abstract} 
The game of Tetris is an important benchmark for research in artificial intelligence and machine learning. This paper provides a historical account of the algorithmic developments in Tetris and discusses open challenges. Handcrafted controllers, genetic algorithms, and reinforcement learning have all contributed to good solutions. However, existing solutions fall far short of what can be achieved by expert players playing without time pressure. Further study of the game has the potential to contribute to important areas of research, including feature discovery, autonomous learning of action hierarchies, and sample-efficient reinforcement learning.

\end{abstract} 

\section{Introduction}
The game of Tetris has been used for more than 20 years as a domain to study sequential decision making under uncertainty. It is generally considered a rather difficult domain. So far, various algorithms have yielded good strategies of play but they have not approached the level of performance reached by expert players playing without time pressure. Further work on the game has the potential to contribute to important topics in artificial intelligence (AI) and machine learning, including feature discovery, autonomous learning of action hierarchies, and sample-efficient reinforcement learning.

 

In this article, we first describe the game and provide a brief history. We then review the various algorithmic approaches taken in the literature. The most recent and successful player was developed using approximate dynamic programming. We conclude with a discussion of current challenges and how existing work on Tetris can inform approaches to other games and to real-world problems. In the Appendix we provide a table of the algorithms reviewed and a description of the features used.


\section{The Game of Tetris}

Tetris is one of the most well liked video games of all time. It was created by Alexey Pajitnov in the USSR in 1984. It quickly turned into a popular culture icon that ignited copyright battles amid the tensions of the final years of the Cold War \cite{docu2004}. 

\begin{figure}[htbp]
   \centering
   \includegraphics[width=0.23\textwidth]{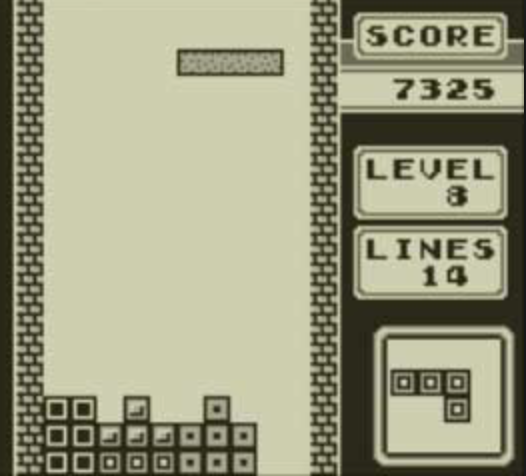} 
    \includegraphics[width=0.23\textwidth]{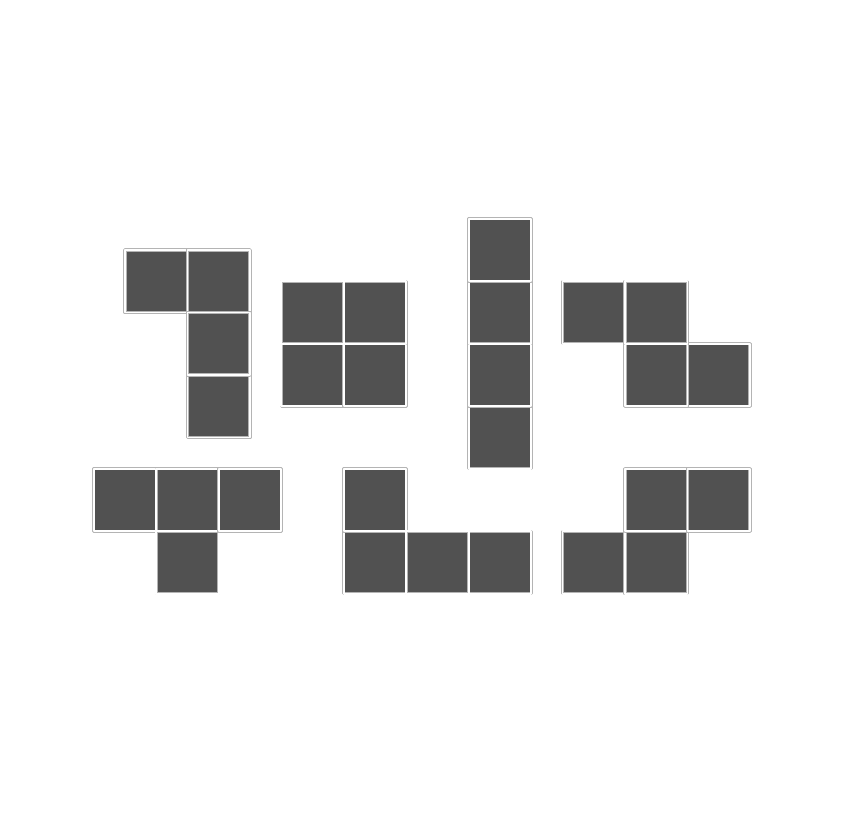} 
   \caption{Screenshot of Tetris on Gameboy and the seven Tetriminos.}
   \label{fig:screenshot}
\end{figure}

The game is played on a two-dimensional grid, initially empty. The grid gradually fills up as pieces of different shapes, called \emph{Tetriminos}, fall from the top, one at a time. The player can control how each Tetrimino lands by rotating it and moving it horizontally, to the left or to the right, any number of times, as it falls one row at a time until one of its cells sits directly on top of a full cell or on the grid floor. When an entire row becomes full, the whole row is deleted, creating additional space on the grid. The game ends when there is no space at the top of the grid for the next Tetrimino. Each game of Tetris ends with probability 1 because there are sequences of Tetriminos that terminate the game, no matter how well they are placed~\cite{Burgiel1997}. Figure \ref{fig:screenshot} shows a screenshot of the game along with the seven Tetriminos.

Despite its simple mechanics, Tetris is a complex game. Even if the full sequence of Tetriminos is known, maximizing the number of rows cleared is NP-complete~\cite{Demaine2003}.

%
%

Tetris can be modeled as a Markov decision process. In the most typical formulation, a state includes the current configuration of the grid as well as the identity of the falling Tetrimino. The available actions are the possible legal placements that can be achieved by rotating and translating the Tetrimino before dropping it. 

This formulation ignores some pieces of information provided in some implementations of the game, for example, the identity of the \textit{next} Tetrimino to fall after the current one is placed. It also excludes actions that are available in some implementations of the game. One example is sliding a Tetrimino under a cliff, which is known as an \textit{overhang}. Another example is rotating the T-shaped Tetrimino to fill an otherwise unreachable slot at the very last moment. This maneuver is known as a \textit{T-spin} and it gives extra points in some implementations of the game.


The original version of the game used a scoring function that awards one point for each cleared line. Subsequent versions allocate more points for clearing more than one line simultaneously. Clearing four lines at once, by placing an I-shaped Tetrimino in a deep well, is allocated the largest number of points. 
Most implementations of Tetris by researchers use the original scoring function, where a single point is allocated to each cleared line.

\section{Algorithms and Features}
\label{sec:algorithms}
Tetris is estimated to have $7 \times 2^{200}$ states. Given this large number, the general approach has been to approximate a value function or learn a policy using a set of features that describe either the current state or the current state--action pair. 

Tetris poses a number of difficulties for research. First, small changes in the implementation of the game cause very large differences in scores. This makes comparison of scores from different research articles difficult. Second, the score obtained in the game has a very large variance. Therefore, a large number of games need to be completed to accurately assess average performance. Furthermore, games can take a very long time to complete. Researchers who have developed algorithms that can play Tetris reasonably well have found themselves waiting for days for a single game to be over.  For that reason, it is common to work with grids smaller than the standard grid size of $20\times10$. A reasonable way to make the game shorter, without compromising its nature, is to use a grid size of $10\times10$. The scores reported in this article are those achieved on the standard grid of size $20\times10$, unless otherwise noted.
 
The most common approach to Tetris has been to develop a linear evaluation function, where each possible placement of the Tetrimino is evaluated to select the placement with the highest value. In the next sections, we discuss the features used in these evaluation functions, as well as how the weights are tuned.

\subsection{Early Attempts}

\citet{tsitsiklis1996feature} used Tetris as a test bed for large-scale feature-based dynamic programming. They used two simple state features: the number of holes, and the height of the highest column. They achieved a score of around 30 cleared lines on a $16\times10$ grid.

\citet{Bertsekas1996} added two sets of features: height of each column, and the difference in height between consecutive columns. Using lambda-policy iteration, they achieved a score of around 2,800 lines. Note, however, that their implementation for ending the game effectively reduces the grid to a size of $19\times10$.

\citet{lagoudakis2002least} added further features, including mean column height and the sum of the differences in consecutive column heights. Using least-squares policy iteration, they achieved an average score of between 1,000 and 3,000 lines. 

\citet{kakade2002natural} used a policy-gradient algorithm to clear 6,800  lines on average using the same features as \citet{Bertsekas1996}.




\citet{farias2006tetris} studied an algorithm that samples constraints in the form of Bellman equations for a linear programming solver. The solver finds a policy that clears around 4,500 lines using the features used by \citet{Bertsekas1996}.

\citet{ramon2004numeric} used relational reinforcement learning with a Gaussian kernel and achieved a score of around 50 cleared lines. \citet{romdhane2008reinforcement} combined reinforcement learning and case-based reasoning using patterns of small parts of the grid. Their scores were also around 50 cleared lines.


\subsection{Hand-Crafted Agent}
Until 2008, the best artificial Tetris player was handcrafted, as reported by \citet{Fahey2003}. Pierre Dellacherie, a self-declared average Tetris player, identified six simple features and tuned the weights by trial and error. These features were number of \textit{holes}, \textit{landing height} of the piece, number of \textit{row transitions} (transitions from a full square to an empty one, or vice versa, examining each row from end to end), number of \textit{column transitions}, \textit{cumulative number of wells}, and \textit{eroded cells} (number of cleared lines multiplied by the number of holes in those lines filled by the present Tetrimino). The evaluation function was as follows:

   $- 4 \times holes - cumulative\:wells \\
   - row\:transitions 
   - column\:transitions \\
   - landing\:height +
    eroded\:cells$

This linear evaluation function cleared an average of 660,000 lines on the full grid. The scores were reported on an implementation where the game was over if the falling Tetrimino had no space to appear on the grid (in the center of the top row). In the simplified implementation used by the approaches discussed earlier, the games would have continued further, until every placement would overflow the grid. Therefore, this report underrates this simple linear rule compared to other algorithms. 


\subsection{Genetic Algorithms}
\citet{bohm2005} used evolutionary algorithms to develop a Tetris controller. In their implementation, the agent knows not only the falling Tetrimino but also the next one. This makes their results incomparable to those achieved on versions with knowledge of only the current Tetrimino. They evolved both a linear and an exponential policy. They reported 480,000,000 lines cleared using the linear function and 34,000,000 using the exponential function, both on the standard grid. They introduced new features such as the number of connected holes, number of occupied cells, and the number of occupied cells weighted by its height. These additional features were not picked up in subsequent research.

\citet{Szita2006} used the cross-entropy algorithm and achieved 350,000 lines cleared. The algorithm probes random parameter vectors in search of the linear policy that maximizes the score. For each parameter vector, a number of games are played. The mean and standard deviation of the best parameter vectors are used to generate a new generation of policies. A constantly decreasing noise allows for an efficient exploration of the parameter space.  Later, \citet{szita2007learning} also successfully applied a version of the cross-entropy algorithm to Ms. Pac-Man, another difficult domain.

Following this work, \citet{Thiery2009, Thiery2009b} added a couple of features (hole depth and rows with holes) and developed the BCTS controller using the cross-entropy algorithm, where BCTS stands for building controllers for Tetris. They achieved an average score of 35,000,000 cleared lines. With the addition of a new feature, pattern diversity, BCTS won the 2008 Reinforcement Learning Competition.

In \citeyear{Boumaza2009}, \citeauthor{Boumaza2009} introduced another evolutionary algorithm to Tetris, the covariance matrix adaptation evolution strategy (CMA-ES). He saw that the resulting weights were very close to Dellacherie's and also cleared 35,000,000 lines on average.

The success of genetic algorithms deserves our attention given the recent resurgence of evolutionary strategies as strong competitors for reinforcement learning algorithms \cite{salimans2017evolution}. Notably, they are easier to parallelize than reinforcement learning algorithms. As discussed below, the latest reported reinforcement learning algorithm uses an evolutionary strategy inside the policy evaluation step \cite{Gabillon2013, Scherrer2015}.

\subsection{Approximate Modified Policy Iteration}

\citet{Gabillon2013} found a vector of weights that achieved 51,000,000 cleared lines using a classification-based policy iteration algorithm inspired by \citet{lagoudakis2003reinforcement}. This is the first reinforcement learning algorithm that has a performance comparable with that of the genetic algorithms. \citeauthor{lagoudakis2003reinforcement}'s idea is to make use of sophisticated classifiers inside the loop of reinforcement learning algorithms to identify good actions. \citet{Gabillon2013} estimated values of state--action pairs using rollouts and then minimized a complex function of these rollouts using the CMA-ES algorithm \cite{Hansen2001}. This is the most widely known evolutionary strategy \cite{salimans2017evolution}. Within this algorithm, CMA-ES performs a cost-sensitive classification task, hence the name of the algorithm: classification-based modified policy iteration (CBMPI). 

The sample of states used by CBMPI was extracted from trajectories played by BCTS and then subsampled to make the grid height distribution more uniform. CBMPI takes at least as long as BCTS to achieve a good level of performance. Furthermore, how best to subsample to achieve good performance is not well understood~\citep[p.~27]{Scherrer2015}.

We next review an approach of a different nature: how the structure of the decision environment allows for domain knowledge to be integrated into the learning algorithm.

\section{Structure of the Decision Environment}
\label{sec:structures}
Real-world problems present regularities of various types. It is therefore reasonable to expect regularities in the decision problems encountered when playing video games such as Tetris. Recent work has identified some of these regularties and has shown that learning algorithms can exploit them to achieve better performance~\cite{Simsek2016}. 

Specifically, \citet{Simsek2016} showed that three types of regularities are prevalent in Tetris: simple dominance, cumulative dominance, and noncompensation. When these conditions hold, a large number of placements can be (correctly) eliminated from consideration even when the exact weights of the evaluation function are unknown. For example, when one placement simply dominates another placement (which means that one placement is superior to the other in at least one feature and inferior in none), the dominated placement can be eliminated.

In Tetris, the median number of possible placements of the falling Tetrimino is 17. \citet{Simsek2016} reduced this number to 3 by using simple dominance to eliminate inferior placements, and to 1 by using cumulative dominance.  




The filtering of actions based on a few indicative features is an ability that can potentially be useful in many unsolved problems. Simpler and more sensitive functions for making decisions can be found after filtering inferior alternatives from the action space, perhaps in a way similar to how  people's attention is guided to a small set of alternatives that they consider worthy.
 

\section{Open Challenges}
\label{sec:challenges}
So far, the transformation of raw squares of the Tetris grid to a handful of useful features has been carried out by hand. Tetris is not yet part of the OpenAI universe or the Atari domain. No deep learning algorithm has learned to play well from raw inputs. \citet{stevensplaying} and \citet{lewis2015generalisation} have reported attempts that achieved at most a couple hundred lines cleared.
%
%

The scoring function where clearing multiple lines at once gives extra points makes a big difference in what policies score well. For this scoring function, it is likely that a linear evaluation function is not the best choice. Tetris would thus constitute a great test bed for learning hierarchies of actions (or options; \citealt{sutton1999between}), where a subgoal could be to set the stage for the I-shaped Tetrimino to clear four lines at once. T-spins are also performed this way: the player needs to set the stage and then wait for the T shape to be able to perform the maneuver. These subgoals are not defined by a unique state but by features of the grid that allow the desired action.
%
%

People enjoy playing Tetris and can learn to play reasonably well after a little practice. Some effort is being made to understand how people do this~\cite{sibert15csc}. Progress in this research may help AI tackle the type of problems that gamers face every day. 

\citet{kirsh1994distinguishing} gave a detailed account of how people perceive a Tetrimino and execute the actions necessary to place it. But an important question is not yet answered: How do people decide where to place the Tetrimino?


Finally, we have not yet developed an efficient learning algorithm for Tetris. Current best approaches require hundreds of thousands of samples, have a noisy learning process, and rely on a prepared sample of states extracted from games played by a policy that already plays well. The challenge is still to develop an algorithm that reliably learns to play using little experience.

\section{Beyond Tetris}
\label{sec:beyond}
Can something be learned from Tetris that can be applied to bigger problems such as real-time strategy (RTS) or open-world games? Tetris may seem like a small game when compared to StarCraft or Minecraft. All of these games, however, share the difficulty that virtually no situation is encountered twice. Furthermore, fast learning should be possible by exploiting the regularities in the game. 

Typically, AI bots facing RTS problems, such as StarCraft, deal with subtasks separately: high-level strategic reasoning is dealt with separately from tactics \citep[p.~4]{ontanon2013survey}. Tetris can be thought of as one of these subtasks, where a simple rule using an appropriate set of features can perform well. Different tasks may need different features. A high-level strategy, for instance, needs spatial features such as \textit{trafficability}\footnote{Trafficability refers to "the ability of a vehicle or unit to move across a specified piece of terrain."} \citep[p.~35]{forbus2002qualitative} whereas tactics, where a unit decides to continue attacking or retreat, may use features such as the number of units. The unit-counting heuristic is reportedly used by a StarCraft bot to decide whether to retreat or keep attacking \citep[p.~10]{ontanon2013survey}. 

Environmental structures such as those found in Tetris are likely to be present in other problems as well. In fact, simple dominance was also observed in backgammon, a complex game requiring multiple high-level strategies to play at expert level~\citep[p.~7]{Simsek2016}. Similar approaches can inform the development of heuristics in problems such as StarCraft. Another example is based on the fact that units on higher ground always have an advantage over units on lower ground~\citep[p~3]{ontanon2013survey}. Strategies where units move to higher ground will likely dominate other strategies. There may be no need to look further.

As AI research continues to deal with increasingly difficult real-world problems, inspiration may come from how people cope with the uncertainties in their lives. Video games, such as Tetris, provide controlled environments where people's decision making can be studied in depth. Game players have limited time and limited computational resources to consider every possible action open to them. Uncertainty about consequences, limited information about alternatives, and the sheer complexity of the problems they face make any exhaustive computation infeasible \citep[p.~169]{simon1972theories}. Yet they make hundreds of decisions in a minute and outperform every existing algorithm in a number of domains. 

\appendix
\section*{Appendix}
\subsection*{Algorithms}

Table \ref{algorithms-table} shows the algorithms that have been used to learn strategies for the game of Tetris, along with their reported scores and used feature sets. 
%
%
\begin{table}[b!]
\vskip 0.15in
\begin{center}
\begin{small}
\begin{sc}
{\setlength{\extrarowheight}{1pt}
\begin{tabular}{L{5cm}L{3.3cm}cL{2.3cm}L{3.2cm}}
\hline
\abovespace\belowspace
  & Algorithm & Grid Size & Lines Cleared & Feature Set Used \\
\hline
\abovespace
\citet{tsitsiklis1996feature} & Approximate value iteration & $16\times10$ & 30 & {Holes and pile height} \\
\citet{Bertsekas1996}  & $\lambda$ - PI & $19\times10$ & 2,800 & {Bertsekas} \\
\citet{lagoudakis2002least}   & Least-squares PI  & $20\times10$ &  $\approx$ 2,000 & {Lagoudakis}\\
\citet{kakade2002natural}    & Natural policy gradient & $20\times10$ & $\approx$ 5,000 & {Bertsekas} \\
Dellacherie \\ {[Reported by \citet{Fahey2003}]}  & Hand tuned & $20\times10$ & 660,000 & {Dellacherie}\\
\citet{ramon2004numeric}     & Relational RL & $20\times10$ & $\approx$ 50 & \\
\citet{bohm2005} & Genetic algorithm & $20\times10$& 480,000,000 (Two Piece)&{B{\"o}hm}\\
\citet{farias2006tetris}      & Linear programming & $20\times10$ & 4,274 & {Bertsekas}\\
\citet{Szita2006} & Cross entropy & $20\times10$ &348,895 & {Dellacherie}\\
\citet{romdhane2008reinforcement} & Case-based reasoning and RL & $20\times10$ & $\approx$ 50 &\\
\citet{Boumaza2009} & CMA-ES & $20\times10$ & 35,000,000& BCTS \\
\citet{Thiery2009, Thiery2009b}& Cross entropy& $20\times10$ & 35,000,000& {DT}\\
\belowspace
\citet{Gabillon2013}& Classification-based policy iteration&$20\times10$& 51,000,000&  {DT} for policy \newline {DT + RBF} for value\\
\hline
\end{tabular}}
\end{sc}
\end{small}
\end{center}
\vskip -0.1in
\caption{Scores and features used by reported algorithms in the game of Tetris. Only the highest score from each publication is reported.}
\label{algorithms-table}
\end{table}

\subsection*{Feature Sets}

{\bf \textsc{Bertsekas:}} Number of \textit{holes}, height of the highest column (also known as \textit{pile height}), \textit{column height}, and \textit{difference in height of consecutive columns}. Twenty-one features in total.

{\bf \textsc{Lagoudakis:}} Number of \textit{holes}, pile height, \textit{sum of differences in height of consecutive columns}, \textit{mean height}, and the change in value of the mentioned features between current and next state. Finally, \textit{cleared lines}. Nine features in total.

{\bf \textsc{Dellacherie:}} number of \textit{holes}, \textit{landing height} of the Tetrimino, number of \textit{row transitions} (transitions from a full square to an empty one, or vice versa, examining each row from end to end), number of \textit{column transitions}, \textit{cumulative wells}, and \textit{eroded cells} (number of cleared lines multiplied by the number of holes in those lines filled by the present Tetrimino). A square that is part of a well is an empty square that can be reached from above with full squares on the sides. A well's depth is the number of vertically connected such squares. \textit{Cumulative wells} is defined as $\sum_w{\sum_{i=1}^{d(w)}{i}}$, where $w$ is a well and $d(w)$ is its depth.

{\bf \textsc{B{\"o}hm:}} \textit{Pile height}, \textit{connected holes} (same as holes but vertically connected holes count as one), \textit{cleared lines}, \textit{difference in height between the highest and lowest column}, \textit{maximum well depth}, \textit{sum of wells' depth}, \textit{landing height} of the Tetrimino, \textit{number of occupied squares}, \textit{number of occupied squares weighted by their height}, \textit{row transitions} and \textit{column transitions}.

{\bf \textsc{BCTS (Building controllers for Tetris):}} Dellacherie's feature set with the addition of \textit{hole depth} (sum of full squares in the column above each hole) and \textit{rows with holes}.  

{\bf \textsc{DT (Dellacherie plus Thiery):}} BCTS's feature set with the addition of \textit{pattern diversity}, which is the number of patterns formed by the difference in height of two subsequent columns. For example, if a column has height 10 and the next one height 9, the pattern is 1. \textit{Pattern diversity} is the number of distinct such patterns with a magnitude lower than 3. 

{\bf \textsc{RBF:}} Radial basis functions of the mean height of the columns. They are defined as $exp(\frac{{-|c-ih/4|}^2}{2(h/5)^2})$, \textit{i} = 0, 1, 2, 3, 4, where \textit{c} is the mean height of the columns and \textit{h} is the total height of the grid.
\clearpage
\bibliography{tetris_refs}
\bibliographystyle{icml2017}

\end{document}